# Tackling Initial Centroid of K-Means with Distance Part (DP-KMeans)

Ahmad Ilham
Informatics Department
Universitas Muhammadiyah Semarang
Semarang, Indonesia
ahmadilham@unimus.ac.id

Danny Ibrahim
Informatics Engineering Department
Universitas Dian Nuswantoro
Semarang, Indonesia
dannyibm10@gmail.com

Luqman Assaffat & Achmad Solichan
Electrical Engineering Department
Universitas Muhammadiyah Semarang
Semarang, Indonesia
{assaffat, solichan}@unimus.ac.id

*Abstract* — **The initial centroid is a fairly challenging problem in the k-means method because it can affect the clustering results. In addition, choosing the starting centroid of the cluster is not always appropriate, especially, when the number of groups increases. The random technique is often used to overcome this problem, but it produces a variety of solutions because the initial centroid initialization uses a random way. Therefore, we propose Distance Part's (DP) method to solve initial cluster initialization problems on the k-means method (DP-KMeans). DP-KMeans is a new approach for initial centroid; this approach works by way of data is partitioned based on the sorted data from largest to smallest value distance to the reference point. This method is called by DP-KMeans, because the data of partition is based on the sorted data distance to the reference point. In this study, four datasets by the UCI machine learning repository are used to evaluate the proposed method. The output process shows that the proposed method produces influential results with the lowest sum of square error for k = 4 are 10.606, 705.144, 13.450, 97.767. Finally, it can be concluded that DP-KMeans can improve the k-means performance on the initial centroid problem.**

## I. INTRODUCTION

Clustering technique is an unsupervised learning used for group objects into groups or clusters so that way the objects in each group share similar resemblances, conversely, each group does not have the same resemblance. The calculation of the distance between data is used to know the resemblance of an object [1]. The smaller distance between the data is the higher similarity between the data. The cluster analysis has been widely applied in various fields such as market research [2], image segmentation [3], pattern recognition [4], decision making and machine learning [5].

Generally, clustering methods are classified into five categories like a partition method, hierarchical method, density-based method and grid method [6]. The most popular clustering methods are partition-based methods and hierarchical methods [7]. The focus in this research is on partition-based clustering method. The partitioning method is divided into two categories of soft partition and hard partition. The partition method separates the data into groups as much as where $k$ where each partition represents cluster [6]. The advantage of partitioning methods shows that it can manipulate large numbers of datasets and takes less computation time than hierarchical methods [8]. Partition methods are grouped into two main parts, hard partition and soft partition. Hard partition grouping of each object must be appropriate in one cluster and not overlap [9]. While the soft partition method, each object has a membership value level in the interval [0, 1] for each cluster [10].

K-means is a partition based clustering method that is often used to categorize data [11]. K-means can be defined as a clustering algorithm that groups data into $k$ clusters based on the closest distance of the data to the cluster center. The advantages of the k-means method are to efficiently group large datasets [12], easy to implemented [11] and a fairly efficient method in terms of time complexity O (nkt) [13]. However, the the k-means method has been limited to numerical data [13] and clustering results depend heavily on the determination initial centroid [14].

There have been many proposed methods for solving initial centroid problems on the k-means method. The most popular methods are MacQueen [15], Al Daoud [16] and Goyal and Kumar [6]. The simplicity of k-means makes this method widely used in many fields.

MacQueen [15] was the first time proposed method by using a random technique for initial centroid on the k-means. The first centroid is randomly selected from the data points. His reason for choosing a centroid randomly is the possibility that the selected centroid is at a dense data point. Then, each data point is grouped by the closest centroid. The new centroid is the average of each group. This process is repeated as long as the cluster data changes. But, there is no mechanism to avoid choosing outliers or centroids that are too close to one another [17]. This method is often considered the same as the Forgy method [18].

In contrast to Al Daoud [16], the proposed method is a variance-based method. This method looks for attributes that have the greatest variant. First, it calculates the variant of each attribute in the dataset, then attributes that have the largest variant select. Furthermore, the attribute with the largest variant is split into k subset, where k is the number of clusters.

The median count of each subset is used for the initialization of the centroid.

Goyal and Kumar [6] also differ from the two methods above. Their proposed Origin Point's method. This method calculates the distance of each data with origin $O$ (0,0), then, the results of distance calculation with origin point are sorted from the smallest to the largest or conversely. After the partition of the data has been sorted into the same partition, hereafter, the average value of each partition is used for the initial centroid initialization on the k-means method. The proposed method yields the same cluster results even though it is run repeatedly. However, this method is still weak which origin point with data is still too far from the centroid cluster, therefore, the initial centroid cluster is less efficient, the results is Sum Square Error (SSE) value is still large.

The method which would be proposed in this research is a new method called Distance Part's (DP) where this method is used for initial cluster initialization on the k-means, thus improving k-means performance by looking at the decrease of sum square error (SSE). DP approach in k-means works by way of data is partitioned based on the sorted data from largest to smallest value distance to the reference point. Called DP because of the data partition based on the sorted of data distance to the reference point in the k-means. The problem formulation in this research is how to improve the performance of the k-means method if DP is used to initialize the initial centroid of the cluster on k-means?

This paper is organized as follows. In section 2, the proposed method is explained. In section 3, the experimental results of comparing the proposed method with others are presented. Finally, our work of this paper is concluded in the last section.

## II. PROPOSED METHOD

We proposed a method called DP-KMeans to tackle initial centroid problem of the k-means method. Distance Part's (DP) method is used for handling the Goyal and Kumar's method [6] weakness in getting the lowest SSE value of the k-means method. The proposed method evaluated using four datasets from the UCI machine learning repository [19]. There are four datasets used like Iris (attribute: 4, number of records: 150), Ionosphere (attribute: 34, number of records: 351), Seeds (attribute: 7, number of records: 210) and User Modeling (attribute: 5, number of records: 258). All dataset used no missing. Fig 1 shows the flowchart of the proposed method.

As it has been shown in Fig 1, there were three improvements made to tackle the Goyal and Kumar's method [6]. Table I showed detail improvement information by our proposed method and Goyal and Kumar's method. In this study, the number of clusters was defined as k=3 and k=4. This below is a detailed explanation of each improvement of the Goyal and Kumar's method with our proposed method:

- *The first improvement*, in Goyal and Kumar's method [6], we conducted some modification in the origin point O (0,0). It was a center origin point which has a value 0, and it was a hint or reference for all of the data which means that the origin point was a center where the attribute value and its record = 0. In our proposed method, the origin point technique was tackled by the reference point average, why? Because the average of centroid data was more available by using the reference point. The reference point average is a mathematical traditional function that commonly used in mathematics exercise, as defined in Eq (1).

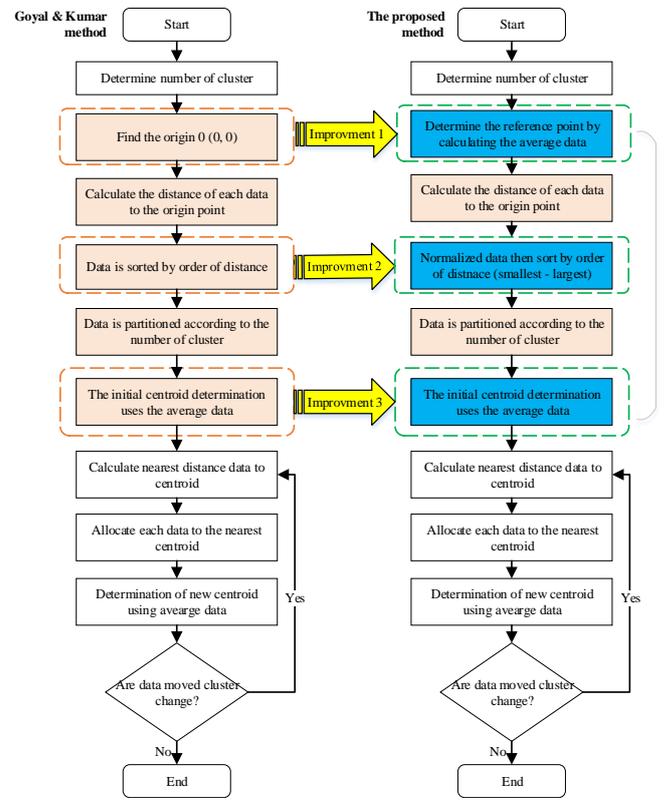

Fig. 1. The proposed method

TABLE I. SUMMARY OF GOYAL & KUMAR METHOD IMPROVED WITH THE PROPOSED METHOD.

| No. | Improved | Goyal & Kumar Method | Proposed Method |
|---|---|---|---|
| 1 | Starting point | Origin point 0 (0, 0) | Average reference point |
| 2 | Sorted data | Original dataset | Data normalized |
| 3 | Initial centroid | Data average | Midrange data |

$$\bar{S}_k = \frac{S_1 + S_2 + S_3 + ... + S_n}{R} \quad (1)$$

where $\bar{S}_k$ is the reference point, $S_n$ is value each of the record data and $R$ is the total number of records.

- *The second improvement*, in Goyal and Kumar's method [6], the sorted data was the original data. But, this technique was weak. Because it divided all of the data which the result of partition didn't define in an optimal partition. Goyal and Kumar's method [6] was repaired by normalizing the original data then sorting it, so that will get the main centroid cluster nearer with the data. In this case, we gave the limit only on real and numeric dataset as defined in Eq (2),

$$D_{Norm} = \frac{D_1, D_2, D_3, ... D_n}{|a|A} \quad (2)$$

Where $D_n$ is value each of data, $|a|$ is the absolute value of data, $A$ is the largest value of the centroid cluster. Then, we sorted from the smallest value to the largest value $Norm_1, Norm_2, Norm_3, ..., Norm_n$. Next, we do sorting distance of each data as defined in Eq. (3),

$$C_n = \frac{D_n + B}{K_n} \quad (3)$$

where $C_n$ is sorting distance each of data record, $B$ is the largest value of the normalized $Norm$ results and $K_n$ is a value each of centroid. In this stage, will be obtained the distance of the cluster center is smaller or close to the data, so the center of the cluster is more homogeneous with the data.

- *The last improvement*, in Goyal and Kumar's method [6], the initial of the centroid used the average of all the data. In our real case, this technique was also weak, because the data was not spreading evenly so the main cluster center was not balanced. Some of the data were too near and the other was too far. The Goyal and Kumar's method [6] was repaired by data midrange technique where the middle value from the minimum and maximum value in the data didn't influence by the distribution or spreading data, so even though the data didn't spread evenly, midrange value would be in the middle between the minimum and maximum data. The data midrange technique was more balance for determining the distance among cluster centers than Goyal and Kumar's method [6] which used average technique among the data. The proposed method was able to increase the performance of the Goyal and Kumar's method [6] of initializing a main centroid of k-means. The data midrange technique could be formulated as follows

$$Me = \frac{a + A}{2} \quad (4)$$

where $a$ is the smallest number and $A$ is the largest number

The detail of the process by the proposed method as follows:

1. Setting up the dataset to use.
2. Determine the number of clusters (K) manually.
3. Centroid initialization.
   a. Determine the reference point ($\bar{S}_k$) by calculating the average value of the data as defined in Eq. (1).
   b. Calculate the distance of each data to the reference point. The distance used is Euclidean, formulated by the following Eq. (2).
   c. Sorting the normalized data based on the smallest distance to the largest. See in Eq. (2).
   d. Once sorted, the data is partitioned by the number of defined clusters of 3 and 4 before, where the first partition sequence is the partition whose members have the closest distance to the reference point, the next partition (the second partition) is the member has a greater distance than the first partition and so on for the following partitions. To calculate the amount of data in a partition by the following Eq. (3).
   e. Calculate the midrange of each data partition as the initial centroid, the calculation of data midrange of each data partition attribute is in Eq. (4).
4. Calculate nearest distance data to centroid. In this study, Euclidean distance was used by following Eq. (5),

$$d_{Euclidean}(x,c) = \sqrt{\sum_{i=1}^{n}(x_i - c_k)^2} \quad (5)$$

where, $x = x_1, x_2, ..., x_n$ and $c = c_1, c_2, ..., c_k$. At this stage, each cluster representation relocated to the centroid of the cluster with the average arithmetic of each cluster. The cause of this method was often called by the cluster mean or centroid like the name they have.

5. Allocate each data to the nearest centroid. At this stage, the data was allocated to the nearest centroid by comparing into each cluster based on the comparison of the distance between the data with each cluster's existing by following Eq. (6).

$$a_{ik} = \begin{cases} 1 & d = \min\{D(xk, vi)\} \\ 0 & other \end{cases} \quad (6)$$

where $a_{ik}$ is a membership of *k*-th data to *i*-th cluster and $vi$ is the value of the centroid *i*-th.

6. Determination of new centroid using average data.
7. Iteration condition.
   When the cluster changed position back to stage 4, otherwise the cluster position didn't change, so the grouping process was complete.

In this study, the proposed method evaluated using the sum squared error (SSE). SSE counting the total number of Errors from the entire data cluster, the smaller the SSE value showed an optimal quality of cluster. Following the Celebi et al [18], SSE defined as (Eq. (7)):

$$SSE \sum_{k=1}^{K} \sum_{x_i \in S_k} \|x_i - c_k\|_2^2 \quad (7)$$

where $\|\cdot\|_2$ denotes the Euclidean norm, $c_k = 1/|S_k| \sum_{x_i \in S_k} x_i$ is centroid on $S_k$.

III. EXPERIMENTAL RESULTS

The experiments were conducted by using a computing platform based on Intel Core i5, 4 GB RAM, and Microsoft Windows 10 64-bit as an operating system and Excel 2013 and

MATLAB as data analytics tools. Excel and MATLAB version R2017a produced a model performance as the calculation output, such as sum square error (SSE) and graphics.

First of all, we conducted an experiment by using only Goyal and Kumar's method [6] without the distance part (DP). The experiment result was taken from Excel and MATLAB as a calculation output shown in Table II. From the method evaluation as shown in Table II and Fig. 2, the sum square error (SSE) of all dataset only k=4 showed an excellent result since was lower rather than k=3 are 83.786, 733.981, 284.032 and 98.704 respectively.

TABLE II. THE METHOD EVALUATION FOR GOYAL & KUMAR'S METHOD [6] ONLY.

| Dataset | Goyal & Kumar's Method [6] | |
|---|---|---|
| | *SSE value for k = 3* | *SSE value for k = 4* |
| Iris | 97.436 | 83.786 |
| Ionosphere | 755.449 | 733.981 |
| Seeds | 313.217 | 284.032 |
| User Modeling | 105.486 | 98.704 |

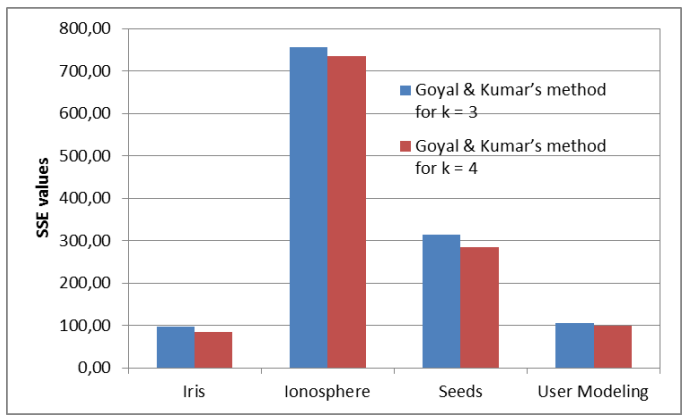

Fig. 2. The method evaluation comparison for Goyal & Kumar's [6] method only.

The second experiment, we implemented Distance Part's (DP) method as the improvement from Goyal and Kumar's method [6] to tackle the initial centroid of k-means. The experimental result as shown in Table III and Fig. 3. In this result for k = 4 of SSE value was more excellent value in all dataset rather than k=3 are 10.606, 705.144, 13.450 and 97.767 respectively.

Finally, we compared the proposed method with other standard k-means and prior research method. We used k=4 to compare all methods because Table II dan Table III show is better than k=3. Table IV shows the comparison between prior research and the proposed method in all dataset which the excellent results comparison was highlighted with boldfaced print.

The proposed method show ed an excellent SSE value of k=3 and k=4 on initialized centroid and outperforms all prior research. As has been shown in Table IV, for k = 3 SSE values in the proposed method were smaller than k-means [15] and Goyal and Kumar's method [6] (there is a decrease in the value of SSE). The same thing happens at k = 4 where the proposed method was superior to the comparison method on all datasets. This showed that the proposed method on each cluster member was more homogeneous than MacQueen [15] and Goyal and Kumar's method [6] in all datasets. However, when compared with the evaluation results, k = 4 had a lowest SSE value more promising than k=3.

The Goyal and Kumar's methods [6] actually have the ability to initialize the initial centroid of a good cluster, this is because the origin point distance to the whole data is still large enough to affect the clustering results. Another weakness was the partition of data which was divided equally based on the number of clusters that have been determined when the optimal is not necessarily divided equally because the number of members of each cluster is not necessarily the same optimal. It had been confirmed in this study that the Goyal and Kumar's method [6] could work well when steps 3, 5 and seven were improved so that the method results in a promising evaluation value rather than the prior research.

TABLE III. THE METHOD EVALUATION FOR IMPROVEMENT GOYAL & KUMAR'S METHOD BY PROPOSED METHOD.

| Dataset | Proposed Method | |
|---|---|---|
| | *SSE value for k = 3* | *SSE value for k = 4* |
| Iris | 12.322 | 10.606 |
| Ionosphere | 754.480 | 705.144 |
| Seeds | 14.788 | 13.450 |
| User Modeling | 105.040 | 97.767 |

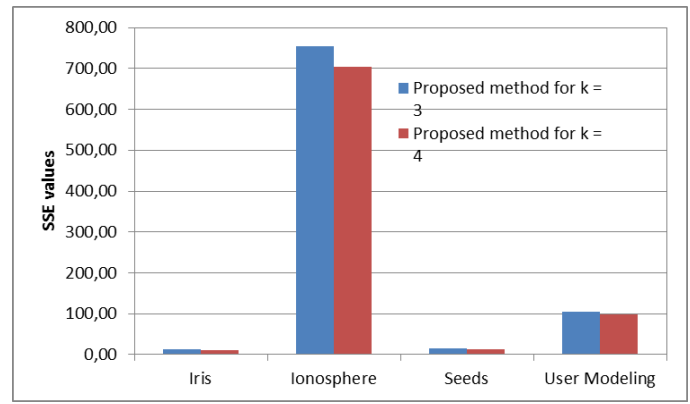

Fig. 3. The method evaluation comparison for the proposed method only.

TABLE IV. COMPARISON TO PRIOR RESEARCH.

| Methods | Years | Results of each dataset | | | |
|---|---|---|---|---|---|
| | | *Iris* | *Ionosphere* | *Seeds* | *User Modeling* |
| MacQueen [15] | 1967 | 84.677 | 746.613 | 313.217 | 98.874 |
| Goyal & Kumar [6] | 2014 | 83.786 | 733.981 | 284.032 | 98.704 |
| Proposed method | 2018 | **10.606** | **705.144** | **13.450** | **97.767** |

IV. CONCLUSION

The results showed that the proposed method yield excellent SSE value especially k=4 because have the lowest value by SSE rather than k=3. The using of DT to improvement Goyal and Kumar's method [6] to initial centroid is proved to increase the performance of k-means. Therefore, it can be concluded that DT able to improve k-means performance for initial centroid.

In this study, there are datasets that have a lot of attributes. Further research can be attributed to the attribute selection method. According to some researchers such as Breaban et al [19] and Tsai et al [20] attribute selection methods can improve the performance of clustering methods by removing irrelevant attributes because not all attributes are informative, so attribute selection methods are very important to use in subsequent research.